\documentclass[runningheads]{llncs}
\usepackage{graphicx}
\usepackage{multirow}
\usepackage{hyperref}
\usepackage{url}
\usepackage{amssymb}
\usepackage{float}
\usepackage{silence}
\usepackage{subcaption}
\usepackage[T1]{fontenc}
\usepackage{xcolor}
\definecolor{light-gray}{gray}{0.95}
\begin{document}
\title{Unsupervised Keyphrase Extraction from
Scientific Publications}
\titlerunning{Unsupervised Keyphrase Extraction from
Scientific Publications}
%
\author{Eirini Papagiannopoulou\inst{1}\orcidID{0000-0002-1014-3481} \and
Grigorios Tsoumakas\inst{1}\orcidID{0000-0002-7879-669X}}

\authorrunning{E. Papagiannopoulou and G. Tsoumakas}
%
\institute{School of Informatics, Aristotle University of Thessaloniki, Greece
\email{\{epapagia,greg\}@csd.auth.gr}}
\maketitle              
\begin{abstract}
We propose a novel unsupervised keyphrase extraction approach that filters candidate keywords using outlier detection. It starts by training word embeddings on the target document to capture semantic regularities among the words. It then uses the minimum covariance determinant estimator to model the distribution of non-keyphrase word vectors, under the assumption that these vectors come from the same distribution, indicative of their irrelevance to the semantics expressed by the dimensions of the learned vector representation. Candidate keyphrases only consist of words that are detected as outliers of this dominant distribution. Empirical results show that our approach outperforms state-of-the-art and recent unsupervised keyphrase extraction methods. 

\end{abstract}
\section{Introduction}
Keyphrase extraction aims at finding a small number of phrases that express the main topics of a document. Automated keyphrase extraction is an important task for managing digital corpora, as keyphrases are useful for summarizing and indexing documents, in support of downstream tasks, such as search, categorization and clustering  \cite{hasan+ng2014}.

We propose a novel unsupervised keyphrase extraction approach based on outlier detection. Our approach starts by learning vector representations of the words in a document via GloVe \cite{Pennington14glove:global} trained solely on this document \cite{papagiannopoulou2018local}. The obtained vector representations encode semantic relationships among words and their dimensions correspond  typically to topics discussed in the document. The key novel intuition in this work is that we expect non-keyphrase word vectors to come from the same multivariate distribution indicative of their irrelevance to these topics. As the bulk of the words in a document are non-keyphrase we propose using the Minimum Covariance Determinant (MCD) estimator \cite{Rousseeuw1984LeastRegression} to model their dominant distribution and consider its outliers as candidate keyphrases. 

\begin{figure}[h]
	\centering
	\includegraphics[width=1.0\linewidth]{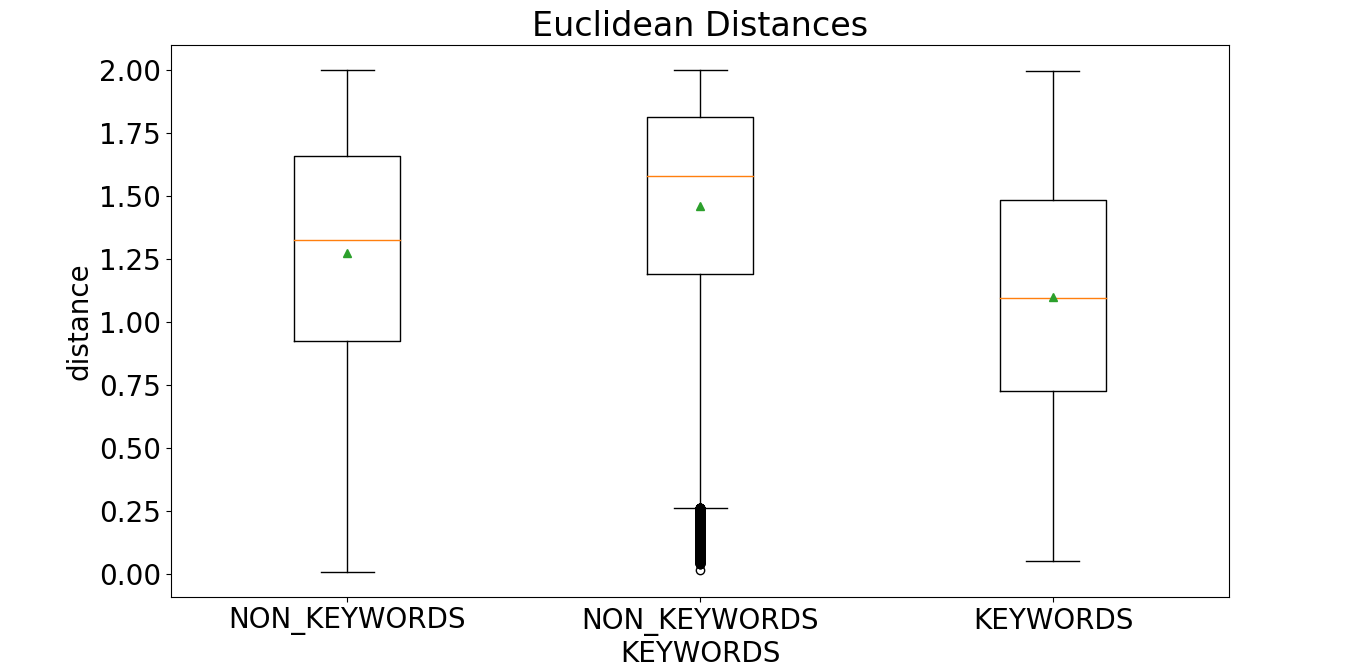}
	\caption{Distribution of Euclidean distances among non-keywords (1st boxplot), between non-keywords and keywords (2nd boxplot), and among keywords (3rd boxplot).}
	\label{fig:euclidean_dist}
\end{figure}

Figure \ref{fig:euclidean_dist} shows the distribution of the Euclidean distances among vectors of non-keyphrase words, between vectors of non-keyphrase and keyphrase words, and among vectors of keyphrase words for a subset of 50 scientific publications from the Nguyen collection  \cite{DBLP:conf/icadl/NguyenK07}. We notice that non-keyphrase vectors are closer together (1st boxplot) as well as the keyphrase vectors between each other (3rd boxplot). However, the interesting part of the figure is the 2nd boxplot where the non-keyphrase vectors appear to be more distant from the keyphrase vectors, which is in line with our intuition.  

\begin{figure}[h]
	\centering
	\includegraphics[width=1.0\linewidth]{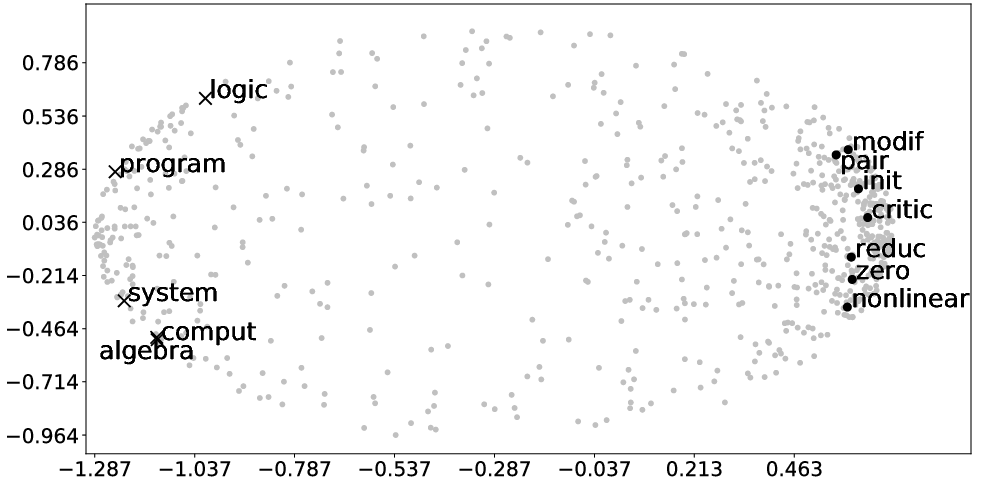}
	\caption{PCA 2d projection of the 5d GloVe vectors in a document. Keyphrase words are the ``x'' in black color, while the rest of the words are the gray circle points. Indicatively, we depict a few non-keywords with black circle points.}
	\label{fig:tsne_findings}
\end{figure}

\noindent Figure \ref{fig:tsne_findings} plots 5d GloVe representations of the words in a computer science article from the Krapivin collection \cite{krapivin2009} on the first two principal components. The article is entitled ``{\em Parallelizing algorithms for symbolic computation using MAPLE}'' and is accompanied by the following two golden keyphrases: logic programming,
computer algebra systems. We notice that keyphrase words are on the far left of the horizontal dimension, while the bulk of the words are on the far right. Similar plots, supportive of our key intuition, are obtained from other documents. 

The rest of the paper is organized as follows. Section \ref{sec:rel_work} gives a review of the
related work in the field of keyphrase extraction as well as a brief overview of
multivariate outlier detection methods. Section \ref{sec:our_approach} presents the proposed keyphrase extraction approach. Section \ref{sec:experiments} describes experimental results highlighting different aspects of our method. We also compare our approach with other state-of-the-art unsupervised keyphrase extraction methods. Finally, Section \ref{sec:conclusions} presents the conclusions and future directions of this work.

\section{Related Work}
\label{sec:rel_work}

In this section, we present the basic unsupervised methodologies (Section \ref{sec:ke_methods}). Then, we briefly review basic multivariate outlier detection methods (Section \ref{sec:outlier_detection_methods}).

\subsection{Keyphrase Extraction}
\label{sec:ke_methods}

Most keyphrase extraction methods have two basic stages: a) the selection of candidate words or phrases, and b) the ranking of these candidates. As far as the first one is concerned, most techniques detect the candidate lexical units or phrases based on grammar rules and syntax patterns \cite{hasan+ng2014}. For the second stage, supervised and unsupervised learning algorithms are employed to rank the candidates. Supervised methods can perform better than unsupervised ones, but demand significant annotation effort. For this reason, unsupervised methods have received more attention from the community. In the rest of this sub-section, we briefly review the literature on unsupervised keyphrase extraction methods.

{\em TextRank} \cite{mihalcea+tatau2004} builds an undirected and unweighted graph of the nouns or adjectives in a document and connects those that co-occur within a window of $W$ words. Then, the PageRank algorithm \cite{grin+page1998} runs until it converges and sorts the nodes by decreasing order. Finally, the top-ranked nodes form the final keyphrases. Extensions to TextRank are {\em SingleRank} \cite{wan+xiao2008} which adds a weight to every edge equal to the number of co-occurrences of the corresponding words, and {\em ExpandRank} \cite{wan+xiao2008} which adds as nodes to the graph the words of the k-nearest neighboring documents of the target document. Additional variations of TextRank are {\em PositionRank} \cite{DBLP:conf/acl/FlorescuC17} that uses a biased PageRank that considers word's positions in the text, and {\em CiteTextRank} \cite{gollapalli2014extracting} that builds a weighted graph considering information from citation contexts. Moreover, in \cite{Wang2014,DBLP:conf/adc/WangLM15} two similar graph-based ranking models are proposed that take into account information from {\em pretrained} word embeddings. In addition, local word embeddings/semantics, i.e.,
embeddings trained from the single document under consideration are used by the Reference Vector Algorithm (RVA) \cite{papagiannopoulou2018local}. RVA computes the mean vector of the words in the document's title and abstract and, then, candidate keyphrases are extracted from the title and abstract, ranked in terms of their cosine similarity with the mean vector, assuming that the closer to the mean vector is a word vector, the more representative is the corresponding word for the publication.

Topic-based clustering methods such as {\em KeyCluster} \cite{liu2009clustering}, {\em Topical PageRank (TPR)} \cite{liu2010automatic}, and {\em TopicRank} \cite{bougouin2013topicrank} aim at extracting keyphrases that cover all the main topics of a document utilizing only nouns and adjectives and forming noun phrases that follow specific patterns. KeyCluster groups candidate words using Wikipedia and text statistics, while TPR utilizes Latent Dirichlet Allocation 
and runs PageRank for every topic changing the PageRank function so as to take into account the word topic distributions. Finally, TopicRank creates clusters of candidates using hierarchical agglomerative clustering. It then builds a graph of topics with weighted edges that consider phrases' offset positions in the text and runs PageRank. A quite similar approach to TopicRank, called MultipartiteRank, has been recently proposed in \cite{Boudin18Multipartite}. Specifically, the incoming edge weights of the nodes are adjusted promoting candidates that appear at the beginning of the document. 

Finally, we should mention the strong baseline approach of \textit{TfIdf} \cite{DBLP:journals/jd/Jones04} that scores the candidate n-grams of a document with respect to their frequency inside the document, multiplied by the inverse of their frequency in a corpus.

\subsection{Multivariate Outlier Detection Methods}
\label{sec:outlier_detection_methods}

Outlier detection methods are categorized into three different groups based on the availability of labels in the dataset \cite{goldstein2016comparative}: \textit{Supervised methods} assume that the dataset is labeled and train a classifier, such as a support vector machine (SVM) \cite{DBLP:journals/sigact/Wille04} or a neural network \cite{DBLP:journals/tnn/Das98}. However, having a labeled dataset of outliers is rare in practice and such datasets are extremely imbalanced causing difficulties to machine learning algorithms. \textit{One-class classification} \cite{DBLP:journals/nn/MoyaH96} assumes that training data consist only of data coming from one class without any outliers. In this case, a model is trained on these data that infers the properties of normal examples. This model can predict which examples are abnormal based on these properties. State-of-the-art algorithms of this category are One-class SVMs \cite{DBLP:journals/neco/ScholkopfPSSW01} and autoencoders \cite{DBLP:conf/dawak/HawkinsHWB02}. The one-class SVM model calculates the support of a distribution by finding areas in the input space where most of the cases lie. In particular, the data are nonlinearly projected into a feature space and are then separated from the origin by the largest possible margin \cite{Dreiseitl2010OutlierDW,DBLP:conf/nips/ScholkopfWSSP99}. The main objective is to find a function that is positive (negative) for regions with high (low) density of points. Finally, \textit{unsupervised methods}, which are the most popular ones, score the data based only on their innate properties. Densities and/or distances are utilized to characterize normal or abnormal cases. 

A popular unsupervised method is Elliptical Envelope \cite{hubert2010minimum,hubert2018minimum,scikit-learn}, which attempts to find an ellipse that contains most of the data. Data outside of the ellipse are considered outliers. The Elliptical Envelope method uses the Fast Minimum Covariance Determinant (MCD) estimator \cite{DBLP:journals/technometrics/RousseeuwD99} to calculate the ellipse's size and shape. The MCD estimator is a highly robust estimator of multivariate location and scatter that can capture correlations between features. Particularly, given a data set $D$, MCD estimates the center, $\bar{x}^*_J$, and the covariance, $S^*_J$, of a subsample $J \subset D$ of size $h$ that minimizes the determinant of the covariance matrix associated to the subsample: 
$$(\bar{x}^*_J, S^*_J): \mathrm{det}~S^*_J \leq \mathrm{det}~S_K\textrm{, }\forall K\subset D, |K|=h$$ 
Another popular unsupervised technique is Isolation Forest (IF) \cite{DBLP:conf/icdm/LiuTZ08}, which builds a set of decision trees and calculates the length of the path needed to isolate an instance in the tree. The key idea is that isolated instances (outliers) will have shorter paths than {\em normal} instances. Finally, the scores of the decision trees are averaged and the method returns which instances are inliers/outliers.

In this work, we are interested in detecting the outliers that do not fit the model well (built by majority of the non-keyphrase words in a text document) or do not belong to the dominant distribution of those words. We expect that the keyphrases and a minority of words that are related to keyphrase words would be the outliers with respect to the dominant non-keyphrase words' distribution or the corresponding model built based on them. 

\section{Our Approach}
\label{sec:our_approach}

Our approach, called {\em Outlying Vectors Rank} (OVR), comprises four steps that are detailed in the following subsections. 

\subsection{Learning Vector Representations}
\label{preprocessing}

Inspired by the graph-based approaches where the vertices added to the graph are restricted with syntactic filters (e.g., selection only nouns and adjectives in order to focus on relations between words of such part-of-speech tags), we remove from the given document all punctuation marks, stopwords and tokens consisting only of digits. In this way, GloVe does not take common/unimportant words into account that are unlikely to be keywords during the model training. Then we apply stemming to reduce the inflected word forms into root forms. We use stemming instead of lemmatization as there are stemmers for various languages. However, we should investigate the possibility of using lemmas, in the future. 

Subsequently we train the GloVe algorithm solely on the resulting document. As training takes place on a single document, we recommend learning a small number of dimensions to avoid overfitting. 
It has been shown in \cite{papagiannopoulou2018local} that such local vectors perform better in keyphrase extraction tasks than global vectors from larger collections. The GloVe model learns vector representations of words such that the dot product of two vectors equals the logarithm of the probability of co-occurrence of the corresponding words \cite{Pennington14glove:global}. At the same time, the statistics of word-word co-occurrence in a text is also the primary source of information for graph-based unsupervised keyphrase extraction methods. In this sense, the employed local training of GloVe on a single document and the graph-based family of methods can be considered as two alternative views of the same information source. 

\subsection{Filtering Non-Keyphrase Words}
\label{sec:outlier_detection}

The obtained vector representation encodes semantic regularities among the document's words. Its dimensions are expected to correspond loosely to the main topics discussed in the document. We hypothesize that the vectors of non-keyphrase words can be modeled with a multivariate distribution indicative of their irrelevance to the document's main topics.  

We employ the fast algorithm of \cite{DBLP:journals/technometrics/RousseeuwD99} for the MCD estimator \cite{Rousseeuw1984LeastRegression} in order to model the dominant distribution of non-keyphrase words. In addition, this step of our approach is used for filtering non-keyphrase words and we are therefore interested in achieving high, if not total, recall of keyphrase words. For the above reasons, we recommend using a quite high (loose) value for the proportion of outliers.
 Then, we apply a second filtering mechanism to the words whose vectors are outliers of the distribution of non-keyphrase words that was modeled with the MCD estimator. Specifically, we remove any words with length less than $3$. We then rank them by increasing position of the first occurrence in the document and consider the top 100 as candidate unigrams, in line with the recent research finding that keyphrases tend to appear closer to the beginning of a document \cite{DBLP:conf/aaai/FlorescuC17}.

Notice that OVR does not have to consider further term frequency thresholds or syntactic information, e.g. part-of-speech filters/patterns, for the candidate keyphrases identification. The properties of the resulting local word vectors capture the essential information based on the flow of speech and presentation of the key-concepts in the article.



\subsection{Generating Candidate Keyphrases}
\label{candidates}

We adopt the paradigm of other keyphrase extraction approaches that extract phrases up to 3 words \cite{Hulth:2003:IAK:1119355.1119383,medelyan2009human} from the original text, as these are indeed the most frequent lengths of keyphrases that characterize documents. As valid punctuation mark for a candidate phrase we consider the hyphen (``-''). Candidate bigrams and trigrams are constructed by considering candidate unigrams (i.e. the top 100 outliers mentioned earlier) that appear consecutively in the document.


\subsection{Scoring Candidate Keyphrases}
\label{scoring}
As a scoring function for candidate unigrams, bigrams, and trigrams we use the TfIdf score of the corresponding n-gram. However, we prioritize to bigrams and trigrams by doubling their TfIdf score, since such phrases are more descriptive and accompany documents as keyphrases more frequently than unigrams \cite{DBLP:conf/ecir/RousseauV15}.

\section{Empirical Study}
\label{sec:experiments}

We first present the setup of our empirical study, including details on the corpora, algorithm implementations, and evaluation frameworks that were used (Section \ref{sec:data_setup}). Then, we study the performance of our approach based on the proportion of the outlier vectors that is considered (Section \ref{sec:results0}), and we compare the performance of the MCD estimator with other outlier detection methods (Section \ref{sec:results1}). In Section \ref{sec:results2}, we compare OVR with other keyphrase extraction methods and we discuss the results. Finally, we give a qualitative (Section \ref{sec:quality}) evaluation of the proposed approach.

\subsection{Experimental Setup}
\label{sec:data_setup}

Our empirical study uses 3 popular collections of scientific publications: a) Krapivin \cite{krapivin2009}, b) Semeval \cite{Kim:semeval2010} and c) Nguyen \cite{DBLP:conf/icadl/NguyenK07}, containing  2304, 244 and 211 articles respectively, along with author- and/or reader-assigned keyphrases. 

We used the implementation of GloVe from Stanford's NLP group\footnote{\url{https://github.com/stanfordnlp/GloVe}}, initialized with default parameters ($x_{max}$ = 100, $\alpha = \frac{3}{4}$, \textit{window size} = 10), as set in the experiments of \cite{Pennington14glove:global}. 
We produce 5-dimensional vectors with 100 iterations. Vectors of higher dimensionality led to worse results. We used the NLTK suite\footnote{\url{https://www.nltk.org/}} for preprocessing.  Moreover, we used the EllipticEnvelope, OneClassSVM, and IsolationForest classes from the scikit-learn library\footnote{\url{https://http://scikit-learn.org}} \cite{Pedregosa2012} for the MCD estimator, One-Class SVM (OC-SVM), and Isolation Forest (IF), respectively, with their default parameters. We utilize the PKE toolkit \cite{DBLP:conf/coling/Boudin16} for the implementations of the other unsupervised keyphrase extraction methods as well as our method. The code for the OVR method will be uploaded to our GitHub repository, in case the paper gets accepted. 

We adopt two different evaluation approaches. The first one is the strict {\em exact match} approach, which computes the $F_1$-score between golden keyphrases and candidate keyphrases, after stemming and removal of punctuation marks, such as dashes and hyphens.  However, we also adopt the more loose {\em word match} approach \cite{DBLP:conf/ecir/RousseauV15}, which calculates the $F_1$-score between the set of words found in all golden keyphrases and the set of words found in all extracted keyphrases after stemming and removal of punctuation marks. We compute $F_1@10$ and $F_1@20$, as the top of the ranking is more important in most applications.

\subsection{Evaluation Based on the Proportion of Outlier Vectors}
\label{sec:results0}

In Tables \ref{tbl:percentage_outliers_exactf1} and \ref{tbl:percentage_outliers_wordf1}, we give the $F_1@10$ and $F_1@20$ of the OVR method using different proportion of outlier vectors, from 10\% up to 49\%, on the three data sets according to the exact match as well as the word match evaluation framework, respectively. Generally, we notice that the higher the outliers' percentage the better is the performance of OVR method. Particularly, in almost all cases (except for the $F_1$@10 of MR based on the word match evaluation), our approach with outlier percentages equal or higher than 30\% outperforms the other competitive keyphrase extraction approaches that their performance is presented in Section \ref{sec:results2} (Tables \ref{tbl:results_exact} and \ref{tbl:results_word}). We set the proportion of outliers for the rest of our experimental study to 0.49 for the Elliptical Envelope method as well as the other outlier detection methods, whose results are given below (Section \ref{sec:results1}), as with this proportion we achieve the highest $F_1$-scores. 

\begin{table}[h]
\caption{$F_1@10$ and $F_1@20$ of the OVR method using different proportion of outlier vectors on the three datasets according to exact match evaluation framework.}
		\label{tbl:percentage_outliers_exactf1}

	\centering
	\scalebox{1.0}{
		\begin{tabular}{|c|c|c|c|c|c|c|}
			\hline
			& \multicolumn{2}{c|}{Semeval} & \multicolumn{2}{c|}{Nguyen} & \multicolumn{2}{c|}{Krapivin}\\ \hline
			\% Outliers    & $F_1$@10 & $F_1$@20 & $F_1$@10 & $F_1$@20 & $F_1$@10 & $F_1$@20 \\ \hline
			10    & 0.130   & 0.124  & 0.164    & 0.139   & 0.113  & 0.086   \\ 
			20    & 0.172   & 0.179  & 0.204    & 0.192   & 0.149  & 0.122   \\ 
            30    & 0.184   & 0.194  & 0.225    & 0.209   & 0.164  & 0.137   \\
            40    & 0.190   & \textbf{0.200}  & 0.230    & 0.212   & 0.169  & 0.143   \\
			49    & \textbf{0.194}   & \textbf{0.200}  &  \textbf{0.237}     & \textbf{0.214}   & \textbf{0.174}  &  \textbf{0.145}  \\ \hline
		\end{tabular}}
			\end{table} 
   
\begin{table}[h]
\caption{$F_1@10$ and $F_1@20$ of the OVR method using different proportion of outlier vectors on the three datasets according to word match evaluation framework.}
		\label{tbl:percentage_outliers_wordf1}
	\centering
	\scalebox{1.0}{
		\begin{tabular}{|c|c|c|c|c|c|c|}
			\hline
			& \multicolumn{2}{c|}{Semeval} & \multicolumn{2}{c|}{Nguyen} & \multicolumn{2}{c|}{Krapivin}\\ \hline
			\% Outliers    & $F_1$@10 & $F_1$@20 & $F_1$@10 & $F_1$@20 & $F_1$@10 & $F_1$@20 \\ \hline
			10    & 0.262   & 0.283  & 0.315    & 0.308   & 0.286  & 0.256   \\ 
			20    & 0.330   & 0.379  & 0.383    & 0.393   & 0.342  & 0.326   \\ 
            30    & 0.349   & 0.408  & 0.414   & 0.426   & 0.369  & 0.353   \\
            40    & 0.358   & 0.417  & 0.425    & 0.435   & 0.384  & 0.346   \\
			49     & \textbf{0.364}   & \textbf{0.424}  &  \textbf{0.433}     & \textbf{0.438}   & \textbf{0.390}  &  \textbf{0.375}  \\ \hline
		\end{tabular}}
		
	\end{table}

\noindent The loose value for the proportion of the outliers helps us in order to apply an effective filtering approach on the candidate keywords that form the keyphrases. We consider that the weak majority of the vocabulary (51\% of inliers) represent a common vocabulary that is used by the author during writing the article, while the strong minority (49\% of outliers) represents the keywords and an accompanying vocabulary that goes hand in hand with the discussion and the description of the keywords (the core concepts of the document). Such information is captured through the co-occurrence statistics, which are utilized by GloVe. 


\subsection{Evaluation Based on the Type of Outlier Detection Method}
\label{sec:results1}

We have designed 2 additional different versions of the proposed OVR approach using 2 alternative outlier detection techniques, which are described previously in Section \ref{sec:outlier_detection_methods}, One-class SVM (OC-SVM) and Isolation Forest (IF). In Tables \ref{tbl:results_outlier_detection_methods_exact} and \ref{tbl:results_outlier_detection_methods_word}, we provide the $F_1@10$ and $F_1@20$ of the different variants of OVR method according to the exact match as well as the word match evaluation framework. Once more, the results confirm that the MCD estimator successfully captures correlations between the vectors' dimensions. 

\begin{table}[h]
\caption{$F_1@10$ and $F_1@20$ of the OVR method using various outlier detection techniques on the three data sets according to exact match evaluation framework.}
		\label{tbl:results_outlier_detection_methods_exact}
	\centering
	\scalebox{1.0}{
		\begin{tabular}{|c|c|c|c|c|c|c|}
			\hline
			& \multicolumn{2}{c|}{Semeval} & \multicolumn{2}{c|}{Nguyen} & \multicolumn{2}{c|}{Krapivin}\\ \hline
			Method    & $F_1$@10 & $F_1$@20 & $F_1$@10 & $F_1$@20 & $F_1$@10 & $F_1$@20 \\ \hline
			OC-SVM    & 0.127   & 0.127  & 0.141    & 0.117   & 0.109  & 0.086   \\ 
			IF        & 0.167   & 0.171  & 0.192    & 0.167   & 0.153  & 0.126   \\ 
			MCD       & \textbf{0.194}   & \textbf{0.200}  &  \textbf{0.237}     & \textbf{0.214}   & \textbf{0.174}  &  \textbf{0.145}  \\ \hline
		\end{tabular}}
		
	\end{table}    
\begin{table}[h]
\caption{$F_1@10$ and $F_1@20$ of the OVR method using various outlier detection techniques on the three data sets according to word match evaluation framework.}
		\label{tbl:results_outlier_detection_methods_word}
	\centering
	\scalebox{1.0}{
		\begin{tabular}{|c|c|c|c|c|c|c|}
			\hline
			& \multicolumn{2}{c|}{Semeval} & \multicolumn{2}{c|}{Nguyen} & \multicolumn{2}{c|}{Krapivin}\\ \hline
			Method    & $F_1$@10 & $F_1$@20 & $F_1$@10 & $F_1$@20 & $F_1$@10 & $F_1$@20 \\ \hline
			OC-SVM    & 0.279   & 0.309  & 0.286    & 0.275   & 0.268  & 0.247   \\ 
			IF        & 0.337   & 0.380  & 0.369    & 0.366   & 0.351  & 0.335   \\ 
			MCD       & \textbf{0.364}   & \textbf{0.424}  &  \textbf{0.433}     & \textbf{0.438}   & \textbf{0.390}  &  \textbf{0.375}  \\ \hline
		\end{tabular}}
		
	\end{table}
\noindent This happens as the classical methods such as OC-SVM can be affected by outliers so strongly that the resulting model cannot finally detect the outlying observations (masking effect) \cite{DBLP:journals/widm/RousseeuwH11}. Moreover, some normal data points may appear as outlying observations. One the other hand, robust statistics, such as the ones that the MCD estimator uses, aim at finding the outliers searching for the model fitted by the majority of the word vectors. Then, the identification of the outliers is defined with respect to their deviation from that robust fit. 

\subsection{Comparison with Other Approaches}
\label{sec:results2}

We compare OVR to the baseline TfIdf method, four state-of-the-art graph-based approaches SingleRank (SR), TopicRank (TR), PositionRank (PR), and MultipartiteRank (MR) with their default parameters, as finally set in the corresponding papers. We also compare our approach to the RVA method which also uses local word embeddings. All methods extract keyphrases from the full-text articles except for RVA which uses the full-text to create the vector representation of the words, but returns keyphrases only from the abstract.

Table \ref{tbl:results_exact} shows that OVR outperforms the other methods in all datasets by a large margin, followed by TfIdf (2nd) and MR (3rd), based on the exact match evaluation framework. TR and PR follow in positions 4 and 5, alternately for the two smaller datasets, but without large differences
between them in Krapivin. RVA achieves generally lower scores according to the exact match evaluation framework, as it extracts keyprases only from the titles/abstracts, which approximately contain half of the gold keyphrases on average \cite{papagiannopoulou2018local}. SR is the worst-performing method in all datasets. 

\begin{table}[h]
\caption{$F_1@10$ and $F_1@20$ of all competing methods on the three data sets according to exact match evaluation framework.}
		\label{tbl:results_exact}
	\centering
	\scalebox{1.0}{
		\begin{tabular}{|c|c|c|c|c|c|c|}
			\hline
			& \multicolumn{2}{c|}{Semeval} & \multicolumn{2}{c|}{Nguyen} & \multicolumn{2}{c|}{Krapivin} \\ \hline 
			Method    & $F_1$@10 & $F_1$@20 & $F_1$@10 & $F_1$@20 & $F_1$@10 & $F_1$@20 \\ \hline 
			SR        & 0.036   & 0.053  & 0.043    & 0.063   & 0.026  & 0.036 \\ 
			TR        & 0.135   & 0.143  & 0.126    & 0.118   & 0.099  & 0.086 \\ 
			PR        & 0.132   & 0.127  & 0.146    & 0.128   & 0.102  & 0.085 \\ 
			MR        & 0.147   & 0.161  & 0.147    & 0.149   & 0.116  & 0.100 \\ 
			TfIdf     & 0.153   & 0.175  & 0.199    & 0.204   & 0.126  & 0.113 \\ 
            RVA		  &	0.094	& 0.124  & 0.097	& 0.114	  &	0.093  & 0.099 \\ 
		OVR       & \textbf{0.194}   & \textbf{0.200}  &  \textbf{0.237}     & \textbf{0.214}   & \textbf{0.174}  &  \textbf{0.145}  \\ \hline
        \end{tabular}}
		
	\end{table}

\begin{table}[h]
\caption{$F_1@10$ and $F_1@20$ of all competing methods on the three data sets according to word match evaluation framework.}
		\label{tbl:results_word}
	\centering
	\scalebox{1.0}{
		\begin{tabular}{|c|c|c|c|c|c|c|}
			\hline
			& \multicolumn{2}{c|}{Semeval} & \multicolumn{2}{c|}{Nguyen} & \multicolumn{2}{c|}{Krapivin} \\ \hline
			Method    & $F_1$@10 & $F_1$@20 & $F_1$@10 & $F_1$@20 & $F_1$@10 & $F_1$@20 \\ \hline 
			SR        & 0.285   & 0.299  & 0.322    & 0.309   & 0.290  & 0.256 \\ 
			TR        & 0.347   & 0.380  & 0.376    & 0.351   & 0.312  & 0.277 \\ 
			PR        & 0.296   & 0.318  & 0.371    & 0.350   & 0.342  & 0.302 \\ 
			MR        & \textbf{0.365}   & 0.403  & 0.407    & 0.383   & 0.342  & 0.303 \\ 
			TfIdf     & 0.308   & 0.368  & 0.370    & 0.394   & 0.309  & 0.305 \\ 
            RVA		  &	0.333	 & 0.366  & 0.374	& 0.379	  &	0.348  & 0.337 \\ 
            OVR       & 0.364   & \textbf{0.424}  &  \textbf{0.433}     & \textbf{0.438}   & \textbf{0.390}  &  \textbf{0.375}  \\ \hline
		\end{tabular}}
	\end{table}

\noindent Moreover, Table \ref{tbl:results_word} confirms the superiority of the proposed method based on the word match evaluation framework. Once more, OVR outperforms the other methods in all datasets by a large margin except for Semeval where MR slightly outperforms OVR.

Based on statistical tests, OVR is significantly better than the rest of the methods in all datasets (besides the MR in Semeval with respect to word match evaluation approach) at the 0.05 significance level. As far as the statistical significance tests concerned, we performed two-sided paired t-test or two-sided Wilcoxon test based on the results of the normality test on the differences of the $F_1$-scores across the three datasets' articles.



\subsection{Qualitative Results}
\label{sec:quality}

In this section, we use OVR to extract the keyphrases of a publication. This scientific article belongs to the Nguyen data collection. We quote the publication's title and abstract below in order to get a sense of its content:
\\\\
{\centering\fcolorbox{black}{light-gray}{\begin{minipage}{33em} \scriptsize{
\textbf{Title: Interestingness of Frequent Itemsets Using Bayesian Networks as Background Knowledge}

\textbf{Abstract:} The paper presents a method for pruning frequent itemsets based on background knowledge represented by a Bayesian network. The interestingness of an itemset is defined as the absolute difference between its support estimated from data and from the Bayesian network. Efficient algorithms are presented for finding interestingness of a collection of frequent itemsets, and for finding all attribute sets with a given minimum interestingness. Practical usefulness of the algorithms and their efficiency have been verified experimentally.

\textbf{Gold Keyphrases:\textit{association rule, frequent itemset, background knowledge, interestingness, Bayesian network, association rules, emerging pattern}}
}
\end{minipage}}}
\\\\

\begin{figure}[h]
\centering
\begin{subfigure}{.496\linewidth}
\centering
\includegraphics[width=1.0\linewidth]{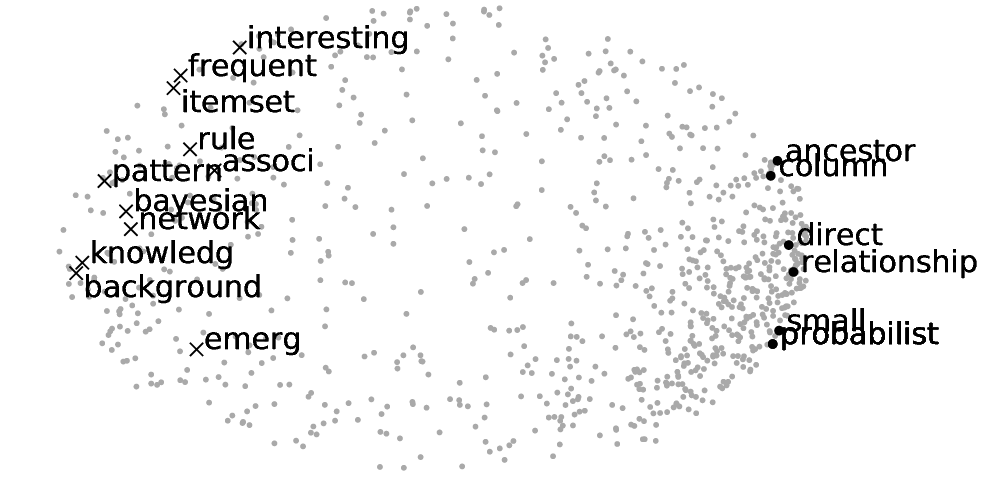}
\caption{}
\label{fig:outliers_inliers}
\end{subfigure}
\begin{subfigure}{.496\linewidth}
\centering
\includegraphics[width=1.0\linewidth]{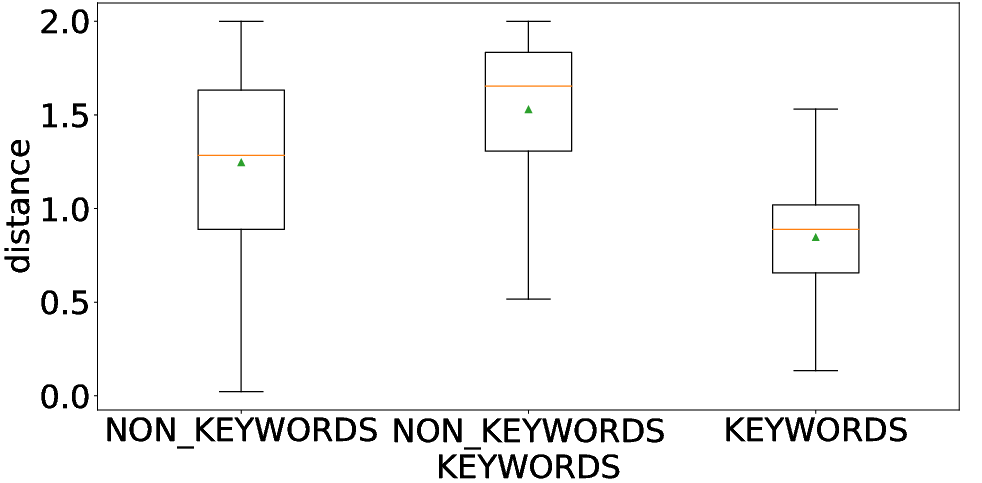}
\caption{}
\label{fig:distances}
\end{subfigure}
\caption{Figure \ref{fig:outliers_inliers} gives the PCA 2d projection of the 5d GloVe vectors of the document, while Fig. \ref{fig:distances} shows the Euclidean distances distribution among non-keywords (1st boxplot), between non-keywords and keywords (2nd boxplot), and among keywords (3rd boxplot)}
\label{fig:statistics}
\end{figure}

In Fig. \ref{fig:statistics}, we give the PCA 2d projection of the 5d GloVe vectors of the document as well as the Euclidean distances distribution among non-keywords, between non-keywords and keywords, and among keywords. Moreover, for evaluation purposes, we transform the set of ``gold'' keyphrases into the following one (after stemming and removal of punctuation marks, such as dashes and hyphens):
\\
{\centering\fcolorbox{black}{light-gray}{\begin{minipage}{33em} \small{{ \textbf{
     \{(associ, rule), (frequent, itemset), (background, knowledg), (interesting), (bayesian, network) (emerg, pattern)\}}
}}
\end{minipage}}}
\\

\noindent The OVR's result set is given in the first box below, by decreasing ranking score, followed by its stemmed version in the second box. The words that are both in the golden set and in the set of our candidates are highlighted with bold typeface:\\


{\centering\fbox{\begin{minipage}{33em} {\small{ \{attribute sets, \textbf{bayesian network}s, \textbf{interestingness}, \textbf{itemset}s, \textbf{background knowledge}, \textbf{bayesian}, attribute, \textbf{frequent itemset}s, interesting attribute, interesting attribute sets, \textbf{interestingness} measure, interesting \textbf{patterns}, \textbf{association rules}, data mining, probability distributions, given minimum, minimum \textbf{interestingness}, given minimum \textbf{interestingness}, minimum support, \textbf{knowledge} represented\} }
}
\end{minipage}}}


{\centering\fbox{\begin{minipage}{33em} {\small{ \{(attribut, set), (\textbf{bayesian}, \textbf{network}), (\textbf{interesting}), (\textbf{itemset}), (\textbf{background, knowledg}), (\textbf{bayesian}), (attribut), (\textbf{frequent}, \textbf{itemset}), (interest, attribut), (interest, attribut, set), (\textbf{interesting}, measur), (interest, \textbf{pattern}), (\textbf{associ}, \textbf{rule}), (data, mine), (probabl, distribut), (given, minimum), (minimum, \textbf{interesting}), (given, minimum, \textbf{interesting}), (minimum, support), (\textbf{knowledg}, repres)\}
}}
\end{minipage}}}
\\\\

According to the exact match evaluation technique, the top-20 returned candidate keyphrases by OVR include 5 True Positives (TPs), 15 False Positives (FPs) and 1 False Negative (FNs). Hence, precision = 0.25, recall = 0.83 and $F_1$ = 0.38. However, according to the word match evaluation technique, the top-20 returned candidate keyphrases by OVR include 10 TPs, 12 FPs and 1 FNs. Hence, precision = 0.45, recall = 0.91, and $F_1$ = 0.60.






\section{Conclusion and Future Work}
\label{sec:conclusions}

We proposed a novel unsupervised method for keyphrase extraction, called Outlying Vectors Rank (OVR). Our method learns vector representations of the words in a target document by locally training GloVe on this document and then filters non-keyphrase words using the MCD estimator to model their distribution. The final candidate keyphrases consist of those lexical units whose vectors are outliers of the non-keyphrase distribution and appear closer to the beginning of the text. Finally, we use TfIdf to rank the candidate keyphrases.

In the next steps of this work, we aim to delve deeper into the local vector representations obtained by our approach and their relationship with keyphrase and non-keyphrase words. We plan to study issues such as the effect of the vector size and the number of iterations for the convergence of the GloVe model, as well as look into alternative vector representations. In addition, we aim to investigate the effectiveness of the Mahalanobis distance in the scoring/ranking process. 


%
%
\bibliographystyle{splncs04}
\bibliography{Mendeley_alma}

\begin{thebibliography}{10}
\providecommand{\url}[1]{\texttt{#1}}
\providecommand{\urlprefix}{URL }
\providecommand{\doi}[1]{https://doi.org/#1}

\bibitem{DBLP:conf/coling/Boudin16}
Boudin, F.: pke: an open source python-based keyphrase extraction toolkit. In:
  Proceedings of the 26th International Conference on Computational
  Linguistics, {COLING} 2016, Proceedings of the Conference System
  Demonstrations. pp. 69--73. Osaka, Japan (December 11-16 2016),
  \url{http://aclweb.org/anthology/C/C16/C16-2015.pdf}

\bibitem{Boudin18Multipartite}
Boudin, F.: Unsupervised keyphrase extraction with multipartite. In:
  Proceedings of the 16th Annual Conference of the North American Chapter of
  the Association for Computational Linguistics Proceedings of NAACL, {NAACL}
  2018. New Orleans (June 1-6 2018)

\bibitem{bougouin2013topicrank}
Bougouin, A., Boudin, F., Daille, B.: Topic{R}ank: Graph-based topic ranking
  for keyphrase extraction. In: Proceedings of the 6th International Joint
  Conference on Natural Language Processing, {IJCNLP} 2013. pp. 543--551.
  Nagoya, Japan (October 14-18 2013),
  \url{http://aclweb.org/anthology/I/I13/I13-1062.pdf}

\bibitem{grin+page1998}
Brin, S., Page, L.: The anatomy of a large-scale hypertextual web search
  engine. Computer Networks  \textbf{30}(1-7),  107--117 (1998).
  \doi{10.1016/S0169-7552(98)00110-X},
  \url{https://doi.org/10.1016/S0169-7552(98)00110-X}

\bibitem{DBLP:journals/tnn/Das98}
Das, S.: Elements of artificial neural networks [book reviews]. {IEEE} Trans.
  Neural Networks  \textbf{9}(1),  234--235 (1998).
  \doi{10.1109/TNN.1998.655048}, \url{https://doi.org/10.1109/TNN.1998.655048}

\bibitem{Dreiseitl2010OutlierDW}
Dreiseitl, S., Osl, M., Scheibb{\"o}ck, C., Binder, M.: Outlier detection with
  one-class svms: An application to melanoma prognosis. AMIA Annual Symposium
  proceedings. AMIA Symposium  \textbf{2010},  172--6 (2010),
  \url{https://www.ncbi.nlm.nih.gov/pmc/articles/PMC3041295/}

\bibitem{DBLP:conf/aaai/FlorescuC17}
Florescu, C., Caragea, C.: A position-biased pagerank algorithm for keyphrase
  extraction. In: Proceedings of the Thirty-First {AAAI} Conference on
  Artificial Intelligence. pp. 4923--4924. San Francisco, California, {USA.}
  (February 4-9 2017),
  \url{http://aaai.org/ocs/index.php/AAAI/AAAI17/paper/view/14377}

\bibitem{DBLP:conf/acl/FlorescuC17}
Florescu, C., Caragea, C.: Position{R}ank: An unsupervised approach to
  keyphrase extraction from scholarly documents. In: Proceedings of the 55th
  Annual Meeting of the Association for Computational Linguistics, {ACL} 2017.
  pp. 1105--1115. Vancouver, Canada (July 30 - August 4 2017).
  \doi{10.18653/v1/P17-1102}, \url{https://doi.org/10.18653/v1/P17-1102}

\bibitem{goldstein2016comparative}
Goldstein, M., Uchida, S.: A comparative evaluation of unsupervised anomaly
  detection algorithms for multivariate data. PloS one  \textbf{11}(4),
  e0152173 (2016)

\bibitem{gollapalli2014extracting}
Gollapalli, S.D., Caragea, C.: Extracting keyphrases from research papers using
  citation networks. In: Proceedings of the 28th {AAAI} Conference on
  Artificial Intelligence. pp. 1629--1635. Qu{\'{e}}bec, Canada (July 27 -31
  2014), \url{http://www.aaai.org/ocs/index.php/AAAI/AAAI14/paper/view/8662}

\bibitem{hasan+ng2014}
Hasan, K.S., Ng, V.: Automatic keyphrase extraction: {A} survey of the state of
  the art. In: Proceedings of the 52nd Annual Meeting of the Association for
  Computational Linguistics, {ACL} 2014, (Volume 1: Long Papers). pp.
  1262--1273. Baltimore, MD, USA (June 22-27 2014),
  \url{http://aclweb.org/anthology/P/P14/P14-1119.pdf}

\bibitem{DBLP:conf/dawak/HawkinsHWB02}
Hawkins, S., He, H., Williams, G.J., Baxter, R.A.: Outlier detection using
  replicator neural networks. In: Data Warehousing and Knowledge Discovery, 4th
  International Conference, DaWaK 2002, Aix-en-Provence, France, September 4-6,
  2002, Proceedings. pp. 170--180 (2002). \doi{10.1007/3-540-46145-0\_17},
  \url{https://doi.org/10.1007/3-540-46145-0\_17}

\bibitem{hubert2010minimum}
Hubert, M., Debruyne, M.: Minimum covariance determinant. Wiley
  interdisciplinary reviews: Computational statistics  \textbf{2}(1),  36--43
  (2010)

\bibitem{hubert2018minimum}
Hubert, M., Debruyne, M., Rousseeuw, P.J.: Minimum covariance determinant and
  extensions. Wiley Interdisciplinary Reviews: Computational Statistics
  \textbf{10}(3),  e1421 (2018)

\bibitem{Hulth:2003:IAK:1119355.1119383}
Hulth, A.: Improved automatic keyword extraction given more linguistic
  knowledge. In: Proceedings of the 2003 Conference on Empirical Methods in
  Natural Language Processing, EMNLP 2003. pp. 216--223. Stroudsburg, PA, USA
  (2003). \doi{10.3115/1119355.1119383}

\bibitem{DBLP:journals/jd/Jones04}
Jones, K.S.: A statistical interpretation of term specificity and its
  application in retrieval. Journal of Documentation  \textbf{28}(1),  11--21
  (1972). \doi{10.1108/00220410410560573},
  \url{https://doi.org/10.1108/00220410410560573}

\bibitem{Kim:semeval2010}
Kim, S.N., Medelyan, O., Kan, M., Baldwin, T.: Semeval-2010 task 5 : Automatic
  keyphrase extraction from scientific articles. In: Proceedings of the 5th
  International Workshop on Semantic Evaluation, SemEval@ACL 2010. pp. 21--26.
  Uppsala, Sweden (July 15-16 2010),
  \url{http://aclweb.org/anthology/S/S10/S10-1004.pdf}

\bibitem{krapivin2009}
Krapivin, M., Autayeu, A., Marchese, M.: Large dataset for keyphrases
  extraction. In: Technical Report DISI-09-055. Trento, Italy (2008)

\bibitem{DBLP:conf/icdm/LiuTZ08}
Liu, F.T., Ting, K.M., Zhou, Z.: Isolation forest. In: Proceedings of the 8th
  {IEEE} International Conference on Data Mining {(ICDM} 2008), December 15-19,
  2008, Pisa, Italy. pp. 413--422 (2008). \doi{10.1109/ICDM.2008.17},
  \url{https://doi.org/10.1109/ICDM.2008.17}

\bibitem{liu2010automatic}
Liu, Z., Huang, W., Zheng, Y., Sun, M.: Automatic keyphrase extraction via
  topic decomposition. In: Proceedings of the 2010 Conference on Empirical
  Methods in Natural Language Processing, {EMNLP} 2010. pp. 366--376.
  Massachussets, USA (October 9-11 2010),
  \url{http://www.aclweb.org/anthology/D10-1036}

\bibitem{liu2009clustering}
Liu, Z., Li, P., Zheng, Y., Sun, M.: Clustering to find exemplar terms for
  keyphrase extraction. In: Proceedings of the 2009 Conference on Empirical
  Methods in Natural Language Processing, {EMNLP} 2009. pp. 257--266. Singapore
  (August 6-7 2009), \url{http://www.aclweb.org/anthology/D09-1027}

\bibitem{medelyan2009human}
Medelyan, O., Frank, E., Witten, I.H.: Human-competitive tagging using
  automatic keyphrase extraction. In: Proceedings of the 2009 Conference on
  Empirical Methods in Natural Language Processing, {EMNLP} 2009. pp.
  1318--1327. Singapore (August 6-7 2009),
  \url{http://www.aclweb.org/anthology/D09-1137}

\bibitem{mihalcea+tatau2004}
Mihalcea, R., Tarau, P.: Text{R}ank: Bringing order into text. In: Proceedings
  of the 2004 Conference on Empirical Methods in Natural Language Processing,
  {EMNLP} 2004. pp. 404--411. Barcelona, Spain (July 25-26 2004),
  \url{http://www.aclweb.org/anthology/W04-3252}

\bibitem{DBLP:journals/nn/MoyaH96}
Moya, M.M., Hush, D.R.: Network constraints and multi-objective optimization
  for one-class classification. Neural Networks  \textbf{9}(3),  463--474
  (1996). \doi{10.1016/0893-6080(95)00120-4},
  \url{https://doi.org/10.1016/0893-6080(95)00120-4}

\bibitem{DBLP:conf/icadl/NguyenK07}
Nguyen, T.D., Kan, M.: Keyphrase extraction in scientific publications. In:
  Proceedings of the Asian Digital Libraries. Looking Back 10 Years and Forging
  New Frontiers, 10th International Conference on Asian Digital Libraries,
  {ICADL} 2007, Hanoi, Vietnam, December 10-13, 2007. pp. 317--326 (2007)

\bibitem{papagiannopoulou2018local}
Papagiannopoulou, E., Tsoumakas, G.: Local word vectors guiding keyphrase
  extraction. Information Processing \& Management  \textbf{54}(6),  888--902
  (2018), \url{https://doi.org/10.1016/j.ipm.2018.06.004}

\bibitem{scikit-learn}
Pedregosa, F., Varoquaux, G., Gramfort, A., Michel, V., Thirion, B., Grisel,
  O., Blondel, M., Prettenhofer, P., Weiss, R., Dubourg, V., Vanderplas, J.,
  Passos, A., Cournapeau, D., Brucher, M., Perrot, M., Duchesnay, E.:
  Scikit-learn: Machine learning in {P}ython. Journal of Machine Learning
  Research  \textbf{12},  2825--2830 (2011)

\bibitem{Pedregosa2012}
Pedregosa, F., Varoquaux, G., Gramfort, A., Michel, V., Thirion, B., Grisel,
  O., Blondel, M., Prettenhofer, P., Weiss, R., Dubourg, V., VanderPlas, J.,
  Passos, A., Cournapeau, D., Brucher, M., Perrot, M., Duchesnay, E.:
  Scikit-learn: Machine learning in python. Journal of Machine Learning
  Research  \textbf{12},  2825--2830 (2011),
  \url{http://dl.acm.org/citation.cfm?id=2078195}

\bibitem{Pennington14glove:global}
Pennington, J., Socher, R., Manning, C.D.: Glove: Global vectors for word
  representation. In: Proceedings of the 2014 Conference on Empirical Methods
  in Natural Language Processing, {EMNLP} 2014. pp. 1532--1543. Doha, Qatar
  (October 25-29 2014), \url{http://aclweb.org/anthology/D/D14/D14-1162.pdf}

\bibitem{DBLP:conf/ecir/RousseauV15}
Rousseau, F., Vazirgiannis, M.: Main core retention on graph-of-words for
  single-document keyword extraction. In: Proceedings of the Advances in
  Information Retrieval - 37th European Conference on {IR} Research, {ECIR}
  2015. pp. 382--393. Vienna, Austria (March 29 - April 2 2015).
  \doi{10.1007/978-3-319-16354-3\_42},
  \url{https://doi.org/10.1007/978-3-319-16354-3\_42}

\bibitem{Rousseeuw1984LeastRegression}
Rousseeuw, P.J.: {Least median of squares regression}. Journal of the American
  Statistical Association  \textbf{79}(388),  871--880 (1984).
  \doi{10.1080/01621459.1984.10477105}

\bibitem{DBLP:journals/technometrics/RousseeuwD99}
Rousseeuw, P.J., van Driessen, K.: A fast algorithm for the minimum covariance
  determinant estimator. Technometrics  \textbf{41}(3),  212--223 (1999).
  \doi{10.1080/00401706.1999.10485670},
  \url{https://doi.org/10.1080/00401706.1999.10485670}

\bibitem{DBLP:journals/widm/RousseeuwH11}
Rousseeuw, P.J., Hubert, M.: Robust statistics for outlier detection. Wiley
  Interdisc. Rew.: Data Mining and Knowledge Discovery  \textbf{1}(1),  73--79
  (2011). \doi{10.1002/widm.2}, \url{https://doi.org/10.1002/widm.2}

\bibitem{DBLP:journals/neco/ScholkopfPSSW01}
Sch{\"{o}}lkopf, B., Platt, J.C., Shawe{-}Taylor, J., Smola, A.J., Williamson,
  R.C.: Estimating the support of a high-dimensional distribution. Neural
  Computation  \textbf{13}(7),  1443--1471 (2001).
  \doi{10.1162/089976601750264965},
  \url{https://doi.org/10.1162/089976601750264965}

\bibitem{DBLP:conf/nips/ScholkopfWSSP99}
Sch{\"{o}}lkopf, B., Williamson, R.C., Smola, A.J., Shawe{-}Taylor, J., Platt,
  J.C.: Support vector method for novelty detection. In: Advances in Neural
  Information Processing Systems 12, {[NIPS} Conference, Denver, Colorado, USA,
  November 29 - December 4, 1999]. pp. 582--588 (1999),
  \url{http://papers.nips.cc/paper/1723-support-vector-method-for-novelty-detection}

\bibitem{wan+xiao2008}
Wan, X., Xiao, J.: Single document keyphrase extraction using neighborhood
  knowledge. In: Proceedings of the 23rd {AAAI} Conference on Artificial
  Intelligence, {AAAI} 2008. pp. 855--860. Chicago, Illinois, USA (July 13-17
  2008), \url{http://www.aaai.org/Library/AAAI/2008/aaai08-136.php}

\bibitem{Wang2014}
Wang, R., Liu, W., McDonald, C.: Corpus-independent generic keyphrase
  extraction using word embedding vectors. In: Software Engineering Research
  Conference (2014)

\bibitem{DBLP:conf/adc/WangLM15}
Wang, R., Liu, W., McDonald, C.: Using word embeddings to enhance keyword
  identification for scientific publications. In: Proceedings of the Databases
  Theory and Applications - 26th Australasian Database Conference, {ADC} 2015.
  pp. 257--268. Melbourne, VIC, Australia (June 4-7 2015).
  \doi{10.1007/978-3-319-19548-3\_21},
  \url{https://doi.org/10.1007/978-3-319-19548-3\_21}

\bibitem{DBLP:journals/sigact/Wille04}
Wille, L.T.: Review of "learning kernel classifiers: Theory and algorithms by
  ralf herbrich." {MIT} press, cambridge, mass., 2002. {ISBN} 026208306x, 384
  pages; and review of "learning with kernels: Support vector machines,
  regularization optimization and beyond by bernhard scholkopf and alexander j.
  smola." {IT} press, cambridge, mass., 2002, {ISBN} 0262194759, 644 pages.
  {SIGACT} News  \textbf{35}(3),  13--17 (2004). \doi{10.1145/1027914.1027921},
  \url{http://doi.acm.org/10.1145/1027914.1027921}

\end{thebibliography}

\end{document}